\pdfoutput=1
\documentclass[11pt]{article}

\newif\ifanonymous
\anonymousfalse %for the non-anonymous version

\ifanonymous
    \usepackage[review]{acl}
\else
    \usepackage{acl}
\fi

\usepackage{times}
\usepackage{latexsym}

\usepackage{booktabs}
\usepackage{multirow}
\usepackage{lipsum}
\usepackage{graphicx}
\usepackage{colortbl}

\usepackage[T1]{fontenc}
\usepackage[utf8]{inputenc}

\usepackage{microtype}

\usepackage{inconsolata}

\title{EmphAssess : a Prosodic Benchmark on Assessing Emphasis Transfer in Speech-to-Speech Models}

\author{
    Maureen de Seyssel\thanks{\hspace{0.5em}Currently at Apple}\hspace{0.3em}\,$^{1,2}$\;\;\; Antony D'Avirro$^{1}$\;\;\; Adina Williams$^{1}$\;\;\; Emmanuel Dupoux$^{1,2}$ \\[1mm]
    $^1$Meta AI Research\;\;\;\;\;\; \\$^2$ENS, EHESS, CNRS, PSL University, France\\[1mm]
    \texttt{maureen.deseyssel@gmail.com} \;\;\;\;\;\;\texttt{\{adavirro,adinawilliams,dpx\}@meta.com}
}

\newcommand{\EC}{\texttt{EmphaClass}}
\begin{document}
\maketitle
\begin{abstract}
We introduce EmphAssess, a prosodic benchmark designed to evaluate the capability of speech-to-speech models to encode and reproduce prosodic emphasis. We apply this to two tasks: speech resynthesis and speech-to-speech translation. In both cases, the benchmark evaluates the ability of the model to encode emphasis in the speech input and accurately reproduce it in the output, potentially across a change of speaker and language. As part of the evaluation pipeline, we introduce \EC, a new model that classifies emphasis at the frame or word level.
\end{abstract}

% \blfootnote{\textcolor{red}{Note : This document remains in a preliminary stage, with particular emphasis on refining sections highlighted in blue. It is subject to further enhancements and modifications.}}
\section{Introduction}

In recent years, significant advancements have been made in the development of Self-Supervised Learning (SSL) models for speech, extending beyond the traditional text-only methods prevalent in the field~\cite{mohamed2022self}. Such speech-based models find successful application across various domains from generative language modelling~\cite{lakhotia2021generative,borsos2023audiolm,nguyen2023generative} to speech-to-speech translation (S2ST)~\cite{jia2019direct,jia2022translatotron,lee2021textless,rubenstein2023audiopalm,barrault2023seamlessm4t}. Unlike text-only models, they exploit additional cues present in the speech signal which are absent in textual input. 

One crucial speech-only cue is prosody. Also termed the ``music of speech'' \citep{wennerstrom2001music}, prosody is marked by the perceived loudness, rhythm, and pitch of speech. Prosody not only adds naturalness to an utterance but also has the capacity to modify the meaning of the conveyed message,  both at a global level, such as in the expression of different emotions, and at a local level, by influencing the interpretation of individual phrases or words~\citep{cutler1997prosody,dahan2015prosody}. For instance, slower speech may suggest hesitation, while altering something like pause placement can actually change the segmentation into words or syntactic constituents, with downstream consequences for the meaning. Hence, accurately capturing these prosodic elements is essential in SSL speech models for any application~\cite{avila2023towards}.

To address this, \citet{kharitonov2021text} proposed explicitly adding prosodically-relevant information such as fundamental frequency and duration to the speech representations models learn, while others aimed at explicitly modelling emotions in such representations~\citep{gan2022iqdubbing, duret2023enhancing}. Although some progress has been made, robust evaluation metrics for prosody remain scarce, and human evaluation, while insightful, is subjective - which can limit reproducibility; as well as being expensive and time intensive - which can hinder its utility in large-scale applications.

Objective evaluations of prosody fall into two main categories: one focuses on utterance-level features like emotion and speech rate to assess \textit{global prosody}, and the other examines \textit{local prosody}, which is concerned with prosodic effects at the level of a word or a phrase, such as breaks, turn ends and emphasis. In addition, one may address prosody for two classes of models: generative decoder-only models (the speech equivalent of GPT \cite{radford2018improving} (e.g. GSLM, \citealp{lakhotia2021generative}; AudioLM, \citealp{borsos2023audiolm}; dGSLM, \cite{nguyen2023generative}), and speech-to-speech (encoder-decoder) approaches, which take speech as input and produce output in a different voice (speech resynthesis) or a different language (S2ST). In this paper, we address the second class of models.

In the context of speech-to-speech (S2S) models, evaluating global prosody can be relatively straightforward, as the features are not directly related to the lexical content.  The assessment of local prosody, however, presents more of a challenge, as it necessitates mapping at the lexical level. This can be relatively feasible in the context of speech resynthesis, where the model directly reconstructs the input signal and, therefore, preserves lexical content (e.g., by correlating prosodic attributes such as duration and fundamental frequency (F0) between input and output utterances; ~\citealp{suni2020prosodic}). However, this becomes more complicated when evaluating S2ST models, as one needs to ensure the correct prosodic feature is applied to the correct word(s)~\cite{duret2023enhancing} (alignment problem).

% Despite progress in creating objective prosody evaluation methods, many remain tailored to specific to particular studies or models. 
Although scarce, there have been recent efforts made to establish benchmarks in the prosodic evaluation of speech models allowing models comparison, including evaluation corpora and pipelines, both at the global prosodic level (pragmatic information : ~\citet{lin2023utility}) and at the local prosodic level (prosodic pauses:~\citet{deseyssel2023prosaudit}). Yet, there is a need for more benchmarks to cover other aspects of prosody, and all types of speech models. 

In this work, we introduce the EmphAssess benchmark, which is focused on local prosody for speech-to-speech models and includes: (i) a new, automatic pipeline for emphasis evaluation that is modular, handles multiple languages and kinds of outputs (including paraphrases and translations, (ii) a novel dataset, the EmphAssess test set, for evaluating model emphasis preservation in English and Spanish according to our pipeline, and (iii) \EC, an emphasis classifier that we finetuned with English data over an existing multilingual SSL model to support our pipeline.

\section{Background}
\paragraph{Emphasis as a prosodic feature.}
Emphasis, the phonetically-realized importance given to particular words or phrases, is critical for interpreting language.  Some of the most important correlates of emphasis are fundamental frequency (f0), duration, and amplitude~\cite{terken2000perception,mo2008acoustic}, although the weight and behaviour of each can vary across languages~\cite{ladd2023prosodic}. These acoustic attributes collectively shape the prosodic contours that signal emphasis in speech. Altering the emphasis in a sentence such as ``I never said he stole my bag" from ``he" to ``stole" can drastically change its meaning. Such nuances are essential for models to process, if they are to have an accurate representation of speech, be they generative language models or S2ST systems. 
In fact, the issue of accurate emphasis transfer in S2ST models has attracted some research attention over the years. Studies by~\citet{tsiartas2013study,do2016preserving,do2018sequence} approach this topic using cascaded models (with separate Automatic Speech Recognition, Machine Translation, and Text-to-Speech models). A more recent approach by~\citet{huang2023holistic}  integrates the two first components into a single encoder module capable of multilingual embeddings. 
% Their Spanish-English model uses global and local prosodic features like F0, duration, and energy to control the output.
Similar to other prosodic features, emphasis in S2S models is primarily evaluated through human evaluation~\cite{tsiartas2013study,huang2023holistic}, although \citet{do2016preserving,do2018sequence} proposed leveraging an emphasis classification algorithm to calculate F1 scores by matching emphasised words in the input and output utterances. Yet, this method is limited to a single language pair and cannot handle variations in translation outputs, only recognising one ``gold'' translation per dataset utterance. Consequently, this metric is ill-suited for comprehensive automatic benchmarking across various models.

\paragraph{Word-level emphasis classification.} As suggested by~\citet {do2016preserving,do2018sequence}, a robust word-level emphasis classification system is critical in automatic evaluation of emphasis transfer in S2ST models. Existing algorithms, predominantly designed for text-to-speech applications, often rely on traditionally engineered features (e.g. MFCCs or Fbanks), sometimes augmented with other prosodic-related information (e.g. F0, duration)~\cite{do2016preserving,heba2017lexical,ning2017learning,zhang2018emphasis}. Some also incorporate lexical information from textual transcripts~\cite{brenier2005detection,zhou2020inferring}. 
However, these models frequently suffer from limited generalisability across different datasets, voice types, and languages. There is a compelling argument for using the speech waveform directly as input to enhance generalisability. To our knowledge, the only study to have adopted this approach is that of \citet{vaidya2022deep}, which employed a CRNN framework for classifying emphasis in children's speech; their work, however, was limited to a single language (and is not open-sourced). We propose that leveraging pretrained models trained on multilingual datasets could result in significant advancements in this field.

% \vspace{1cm}
% Consequently, our study seeks to fill an existing gap by proposing an evaluation benchmark for S2S emphasis that is translation-independent, model-agnostic, objective, and universally applicable across language pairs.

\begin{figure*}[htpb!]
  \centering
\includegraphics[width=\linewidth]{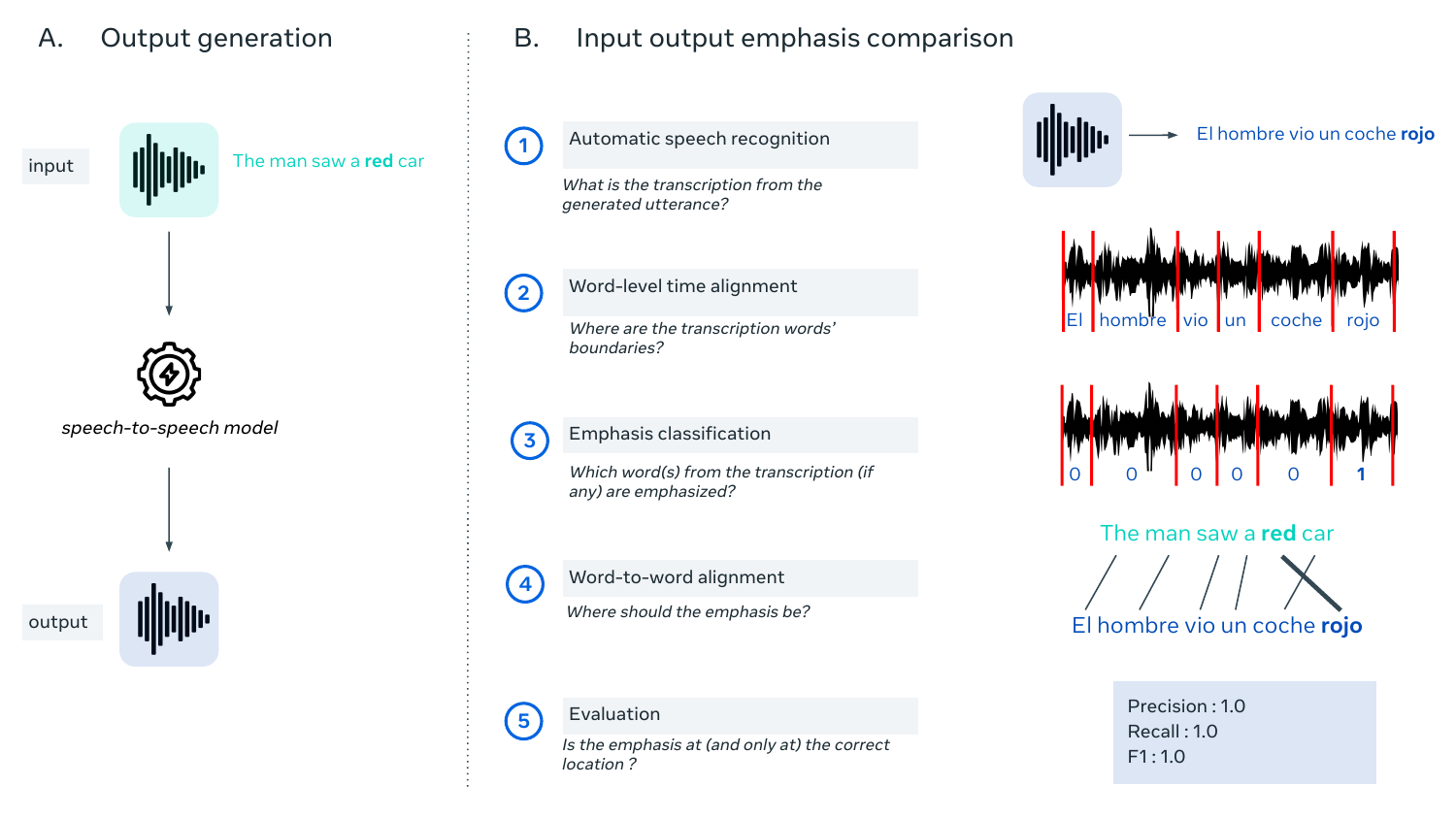}
  \caption{\textbf{Overview of the EmphAssess evaluation pipeline.} Left panel : Output generation. Right panel : Input-output emphasis comparison.}
\label{fig:evaluation_pipeline}
\end{figure*}

\section{Introducing EmphAssess}

In this study, we introduce EmphAssess, a versatile automatic benchmark for evaluating emphasis preservation in S2S models, including S2ST ones. Essentially, this benchmark comprises a carefully curated dataset of English utterances with emphasised words, accompanied by an automatic evaluation pipeline, and results on some of the most recent S2S SSL models. Our evaluation framework, inspired by the methodology of \citet{do2016preserving,do2018sequence}, assesses emphasis alignment between the source and the model's output utterances. Our benchmark's novelty lies in its capacity to handle various output types, including paraphrases and translations. 

Guided by the data we have for setting optimal baselines, the EmphAssess benchmark is specifically designed for English-to-English and English-to-Spanish S2S models. However, our work goes further, laying the groundwork for extending this benchmark to other language pairs. Moreover, the evaluation pipeline itself is already capable of being applied to a broad spectrum of language pairs. Also, while we focus here on unsupervised speech language models, EmphAssess is versatile enough to be applied to any S2S framework.

The EmphAssess evaluation pipeline's modular structure is a key feature, with each module designed to function independently and allow for straightforward modifications. We leverage a suite of distinct open-source models, each finetuned for particular tasks. The pipeline can therefore be upgraded to incorporate improvements in each module seamlessly.  Although such enhancements may necessitate a re-evaluation of the models within our benchmark, this inherent adaptability is a considerable benefit, ensuring EmphAsses can remain current with the latest research for years to come.

Finally, we introduce and open-source, as part of this automatic evaluation pipeline, a novel emphasis classifier at the word level: \EC. This classifier is finetuned over an existing multilingual SSL model with the hope of enhancing its robustness across multiple languages and variability.

The evaluation code, emphasis classifier and dataset introduced in this paper are available in our related repository
\ifanonymous
    \footnote{Final repository will be made available after the double-blind review.}.
\else
    \footnote{https://github.com/facebookresearch/emphassess}.
\fi

%NOTE : ADD INFO AS THE FACT THA ALL COD IS AVAILAB

\section{The EmphAssess Dataset}

The EmphAssess dataset comprises synthetically generated speech utterances, each containing at least one emphasised word. Accompanying these utterances are metadata detailing the transcription, the positional index of the emphasised word(s), and information about the synthetic voice employed for synthesis. In total, the dataset boasts 3652 speech samples derived from 913 unique transcripts (with each transcript being rendered in 4 distinct voices). 
% The dataset is publicly accessible \textbf{\textcolor{red}{at the provided address}}.

The dataset generation started with a selection of transcripts from a list of handwritten transcripts with emphasis annotations\footnote{The emphasis could be applied to any sentence constituents, but it followed a contrastive pattern.} previously created for company-internal Text-to-Speech purposes.
Transcripts containing characters beyond letters or specific punctuation marks\footnote{Retained punctuation characters include: [,:;.?!()]} or those featuring proper nouns (identified using the NLTK toolkit; \citealt{bird2006nltk}) were excluded, to ensure the translations are as straightforward as possible. Moreover, we ensured a minimum of two distinct versions with different emphases for string identical sentences (those with matching word tokens but possibly differing emphasis position indices). This approach was adopted to mitigate any bias should a model exhibit a preference for emphasising a particular word over others.
Finally, we filtered out transcripts that could face alignment challenges with emphasised words during translation. We set up an algorithm to assess the difficulty of aligning emphasised words in an English sentence with their counterparts in multiple target languages, using the SimAlign word-alignment tool~\cite{sabet2020simalign}. Simply put, if an emphasised word in the source matched consistently to a corresponding word across a list of other languages (German, French, Spanish, and Chinese), the sentence was labelled ``easy''; otherwise, it was deemed ``difficult.'' Only ``easy'' transcripts were retained for our dataset.
We were left with 913 distinct transcriptions (with varying emphases) derived from a pool of 299 unique transcriptions. We ensured that the distribution of transcripts was well balanced, in terms of where the emphasis was located.

Next, we employed an internal Text-to-Speech (TTS) tool with a 16 kHz sample rate to synthesise all 913 transcripts, each in the four distinct open-source English Expresso voices \cite{nguyen2023expresso}, namely \texttt{ex01}, \texttt{ex02}, \texttt{ex03} and \texttt{ex04}, resulting in a comprehensive set of 3,652 speech samples.

Finally, we compiled a dataset that is available as part of the benchmark. This dataset comprised four columns: an \texttt{id} column that denotes the unique identifier for each speech segment, a \texttt{src\_sentence} column that contains the corresponding tokenised text transcript presented in list format, a \texttt{gold\_emphasis} column that highlights the index of the emphasised word(s) also in list format, and a \texttt{voice} column that specifies the particular Expresso voice employed for the synthesis. 

\section{The EmphaAssess Evaluation Pipeline}

The evaluation pipeline, as illustrated in Figure \ref{fig:evaluation_pipeline}, is divided into two main stages. The first one (left panel) corresponds to the generation of utterances from the evaluated S2S model. That is, for each utterance from the EmphAssess dataset, we need to generate the corresponding utterance output from the evaluated model. Hence, this inference stage is dependent on the model tested, and we will not expand on it here.

In the second stage (right panel), we perform the automatic evaluation by comparing the input and output utterances. The objective is twofold: firstly, to ascertain whether the emphasis is \textit{retained} in the generated utterance, and secondly, to determine whether the emphasis is \textit{correctly positioned} on the corresponding word. At this stage, available resources include the input (original) utterance, the corresponding output utterance, and the tokenised transcript of the input with the location of the emphasised word(s) identified.
A schematic overview of the evaluation pipeline is shown in the right panel of Figure \ref{fig:evaluation_pipeline}. Initially, we obtain a transcription of the generated utterance \textit{(1)} and the time-aligned word boundaries \textit{(2)}. This information can be used in addition to the raw waveform to detect emphasis at the word level in the output utterance using a classifier \textit{(3)}. At this stage,  we must determine which word(s) in the generated utterance should be emphasised to obtain evaluation scores \textit{(4)}. We use word-to-word alignment at the text level to address this, a technique borrowed from the machine translation field.
Finally, we can use this information to compute precision, recall and F1 score \textit{(5)}. We will now detail our methodology for each of these steps. 

\subsection{Automatic speech Recognition and word-level forced time-alignment}

To achieve accurate transcription of the generated utterance and its associated word-level time-alignments, we utilise the WhisperX system \cite{bain2023whisperx}. This system, which relies on the weakly supervised speech recognition model Whisper~\cite{radford2023robust} for speech transcription, allows retrieval of accurate word-level timestamps, in a variety of languages. 

% Note : I dont think its worth mentioning which whisper model we use knowing that we will have the info in the repo. 

\subsection{Word Emphasis Classification}

As the next step requires detecting emphasis at the word level from the waveform and its corresponding transcription, we propose \EC, a new model for emphasis classification. Our approach was centred around finetuning a pretrained SSL speech model through a frame-classification task to classify a frame as either emphasised or not. We can then aggregate frame-level scores to derive word-level emphasis classifications.

\begin{figure*}[htpb!]
  \centering
  \includegraphics[width=\textwidth]{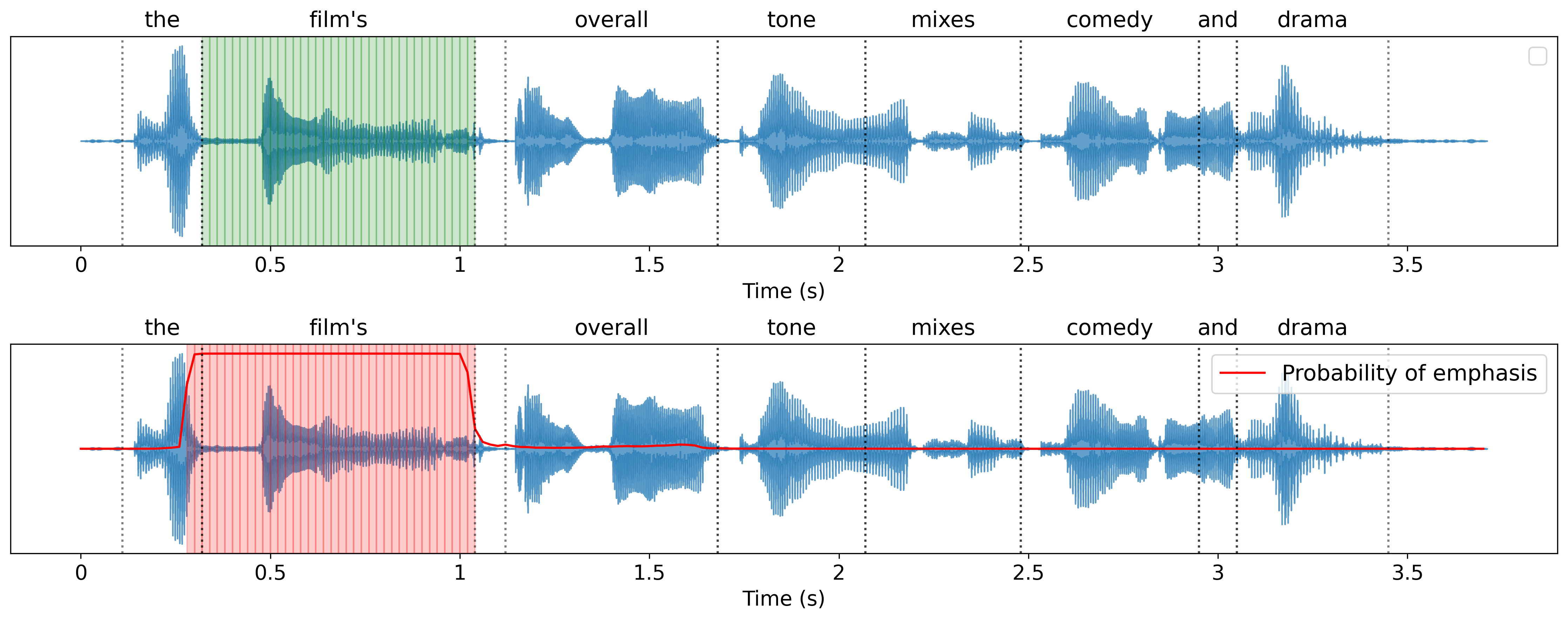}
  \caption{\textbf{Illustrative example of emphasis classification with the trained classifier.} Top: gold annotations. Bottom: Emphasis classifier predictions.}
  \label{fig:example_class}
\end{figure*}

\vspace{0.7em}
\noindent\textbf{Data.}
We utilised speech sourced from the English Expressive Expresso dataset \cite{nguyen2023expresso}. Indeed, this dataset comprises utterances that contain emphasised words, accompanied by their annotations, presented in a diverse range of speaking styles. We retained only those utterances that had at least one word emphasised. We divided the four speakers into two for validation (\texttt{ex03} and \texttt{ex04}) and two for the test set (\texttt{ex01} and \texttt{ex02}). Additionally, we had utterances from six other speakers recorded under identical conditions and with similar emphasis annotations. These were utilised to create an internal training set, amounting to 2.06 hours of speech.
% It is worth noting that these utterances can represent various speaking styles found in the Expresso dataset, including laughter, whispering, and confusion. 
We then used the Montreal Forced Aligner to align the transcription with the audio and obtain reliable word boundaries \cite{mcauliffe2017montreal}. We subsequently processed the data to provide annotations at the frame level regarding emphasis. We deem a frame as `emphasised' if it falls within a word annotated as such, with each frame corresponding to 20ms of speech.
% Our experiments leveraged the comprehensive English Expresso dataset \misscite. From this dataset, four open-source voices were reserved for validation and testing, with the remainder used for training. We specifically utilized transcripts that contained emphasis annotations. However, these transcripts were not restricted to a sole emphasis speaking style, allowing for variations like laughter, whispering, or confusion, as long as emphasis on words was present. For data preprocessing, a forced alignment between the transcription and the waveform was executed to ascertain word boundaries. This subsequently enabled frame-level emphasis annotations.

% Given access to the Spanish counterpart of the Expresso dataset, we processed this data identically, aiming to explore the cross-language generalization capabilities of our model.

\vspace{0.7em}
\noindent\textbf{Emphasis classifier architecture.} We finetuned the multilingual SSL speech model, XLS-R \cite{babu2021xls}, grounded in the Wav2Vec 2.0 architecture \cite{baevski2020wav2vec}. This finetuning encompassed a binary frame classification task using cross-entropy loss, and was carried out using the Wav2Vec2ForAudioFrameClassification method from HuggingFace \citep{wolf2019huggingface}. Our choice of the XLS-R model for extended training and evaluations stemmed from its exceptional performance metrics and promising potential for cross-language generalisation. 

% An exhaustive hyperparameter search was undertaken using Weights and Biases \misscite, with a focus on optimising the F1 score. 

\vspace{0.7em}
\noindent\textbf{Evaluation.}  We use F1 score as the primary metric for evaluating our emphasis classifier, both at the frame and word level. For word-level classification, we compute the average accuracy of the frames within the boundaries of each word. A word was deemed emphasised if more than 50\% of its frames were classified as such. A representative example of this classification is illustrated in Figure \ref{fig:example_class}.

We evaluate the classifier on our test set split of the Expresso dataset, but also on the utterances used in our EmphAssess dataset. Results are presented in Table \ref{Tab:EmphClassResults}. The scores suggest that the model performs well at classifying emphasis in both the Expresso dataset 78.4\% and the Emphasses dataset 93.48\%. The lower scores from the Expresso dataset, compared to the EmphAssess dataset, can be attributed to two factors. Firstly, the Expresso dataset incorporates utterances with speaking styles where the emphasis is notably challenging to discern, such as whispering and laughing. Secondly, using synthetic voices in EmphAssess might offer more consistent and clearer patterns of emphasis than the natural utterances from Expresso, making it easier for the classifier to discern, and thus leading to higher accuracy scores. 
% We present more visualisations of the classification resulting from the classifier in \textcolor{red}{our online demo.}

\begin{table}[htpb]
\resizebox{\columnwidth}{!}{

\begin{tabular}{@{}llrrrrrrrr@{}}
\toprule
\multirow{2}{*}{Test data} & \multicolumn{3}{l}{\textbf{Frame-level (\%)}}                                                                                                             & \multicolumn{3}{l}{\textbf{Word-level(\%)}}                                                                                                               \\
                           &                      \multicolumn{1}{l}{\textbf{F1}} &\multicolumn{1}{l}{\textbf{Prec.}} & \multicolumn{1}{l}{\textbf{Rec.}} & \multicolumn{1}{l}{\textbf{F1}} & \multicolumn{1}{l}{\textbf{Prec.}} & \multicolumn{1}{l}{\textbf{Rec.}} \\ \midrule
EmphAssess                                        & 89.77                                 & 89.71                                  & 91.72                               & 93.48                                                              & 93.81                                  & 94.04                               \\
Expresso EN                                   & 75.52                                & 60.82                                  & 76.90                                & 78.40                                                            & 56.93                                  & 76.90                                     \\  \bottomrule

\end{tabular}
}
\caption{Results of \EC on The EmphAssess dataset and a subset of the Expresso dataset. F1 score, precision and recall}
\label{Tab:EmphClassResults}
\end{table}

We also ran cross-languages analyses, testing the model on other languages, which results showed that the model can, to some extent, classify other languages. This suggests our research may have utility beyond just the English and Spanish languages we explicitly support. More information is presented in Appendix \ref{app:cross-lang-classifier}.

% \vspace{1em}
% \noindent\textbf{Cross-language analysis.} In order to evaluate the loss of a model performance when tersted on another language - as will  potentially be the case in the scope of our evaluation - we leveraged the existence of the Spamsh expresso to train two other classifiers on top of the existing english one, on eon the Spanish set and on eon the combination o the training an Sapnish test sets. It must be noted that the spansih auqlity being of less good quality , we do not necessarily expect xxx. 

\subsection{Word-to-word alignment}

Returning to the automatic emphasis evaluation pipeline,  we can detect which word(s) is emphasised in an output utterance with the classifier described above, given a waveform, its transcriptions and word boundaries. At this point, we need to identify which word(s) should be emphasised in the output utterance to compute a score for the quality of emphasis transfer. This step is vital because it lets us evaluate any output utterance, including paraphrases and translations, without being limited to a ``gold'' output. To do this, we use a word-to-word alignment algorithm, often seen in machine translation, especially the SimAlign one~\cite{sabet2020simalign}. This tool can align words between two text sentences. Although typically used in machine translation, it's also effective for paraphrases in the same language. A key benefit of SimAlign is that it works across many languages without requiring finetuning. For our needs, we compare the original text input with the output utterance transcription from the ASR to see which word(s) match the emphasised word in the original sentence.

\begin{figure*}[htpb]
  \centering
  \includegraphics[clip, trim=0cm 6.6cm 0cm 0cm, width=1.00\textwidth]{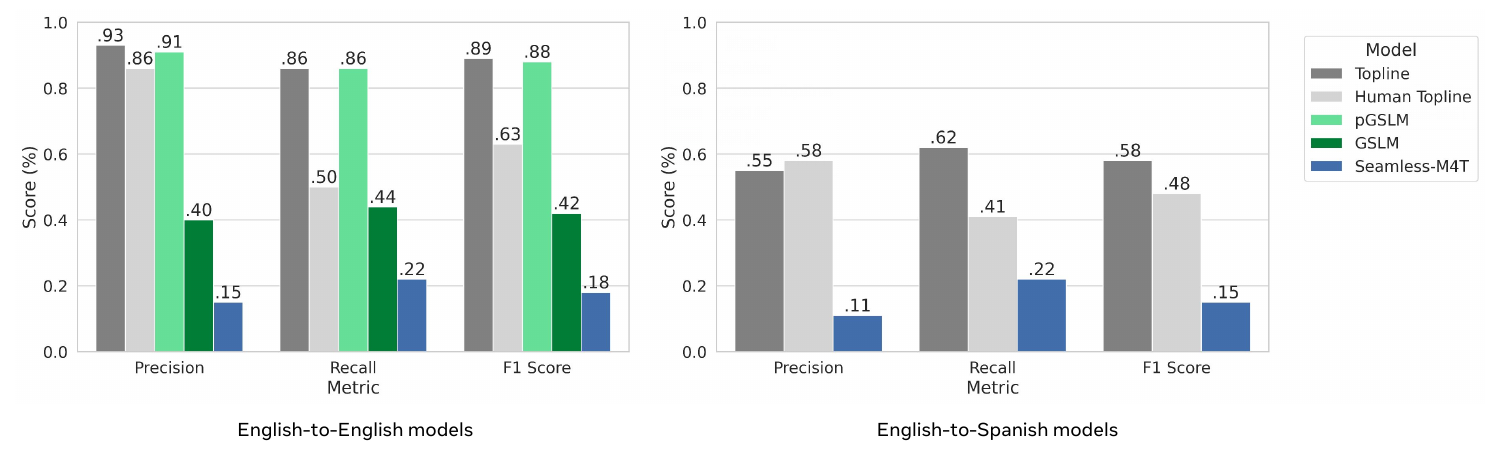}
  \caption{\textbf{Precision, recall and F1 scores on the EmphAssess benchmark.} Left : English-to-English models and English Emphasis classifier. Right : English-to-Spanish models and Spanish Emphasis classifier. }
  \label{fig:en-en-results}
\end{figure*}

% \begin{figure*}[htpb]
%   \centering
%   \includegraphics[width=0.85\linewidth]{images/en-en_enclass_human.png}
%   \caption{Precision, recall and F1 scores on the English-to-English models, using the English Emphasis Classifier.}
%   \label{fig:en-en-results}
% \end{figure*}

\subsection{Metrics}

In the final step, we compare the words that were meant to be emphasised (from the previous step) with the words that were actually emphasised (from the emphasis classification phase). By doing this comparison, we can determine precision, recall, and F1 scores for the whole dataset. 

% The evaluation pipeline is entirely accessible\textbf{ \textcolor{red}{at the following address}.}

\section{Results}

We benchmarked a series of models on the EmphAssess evaluation, both within language (English to English) and using translations (English to Spanish).

\subsection{English S2S models}
We first present results on models that generate speech with the target and source language being identical, here English (left panel of Figure \ref{fig:en-en-results}). This encompasses models that undergo an encoding-decoding method, simply resynthesising the learnt units and those which can learn paraphrases. 

For a topline evaluation, we matched the input utterances from EmphAssess with themselves (that is, we pretended the output utterances were the same as the input ones). This gave us an insight into the best achievable scores, with any potential loss in performance due to problems in the dataset or the various comparison stages. This topline produced an F1 score of 89\%, indicating that our cascaded pipeline performs well.
It should also be noted that we consider chance-level to yield scores of 0, corresponding to a model which does not encode emphasis and thus should not produce any emphasis.

% We alsi introduced two baselines for comparison. The "no-emphasis" baseline assumes no emphasis is produced anywhere, yielding scores of 0. In the "random emphasis" baseline, we simulate randomly applying emphasis to words in the output, matching the number of emphasised words from the input. After testing this over 1000 iterations, we got a consistent score of about 0.147 for F1, precision, and recall (the scores stay the same due to the steady number of emphasised words despite their random placement).However, this method has limitations, such as not considering one input word mapping to multiple output words, and wrongly suggesting all S2S models produce emphasis. Hence, the "no emphasis" baseline with a 0 score is more accurate.

We first assessed the generative GSLM model \cite{nguyen2023generative}, specifically the HuBert, 100 units version. This model initially encodes speech into continuous forms using HuBert \cite{hsu2021hubert}, which are then quantised into units for language modelling. Subsequently, a synthesiser converts these units back to speech. In our study, we extracted the quantised representations from our EmphAssess dataset's speech samples and directly resynthesised them, bypassing the generative language modelling phase. Despite scoring notably lower than the topline with an F1 of 42\%, the model successfully transferred some emphasis to the output utterances. This indicates the presence of prosodic information within these units learned from SSL speech model, a finding supported by \citet{deseyssel2022probing,deseyssel2023prosaudit}.

We also assessed the pGSLM variant, which incorporates extra prosodic features during training to enhance prosody modelling \cite{kharitonov2021text}\footnote{We opted for the variant with continuous input and shift, as it was the top performer in \citet{deseyssel2023prosaudit}.}. Notably, the pGSLM models achieved scores close to the topline, with an F1 of 88\%, highlighting their excellent proficiency in encoding emphasis accurately.

Finally, we assessed the Seamless M4T model \cite{barrault2023seamlessm4t}, forcing it to generate outputs in English. Contrary to the previous models, which generate output constrained in their lexical input, this one is primarily a S2ST model and can output paraphrases. We did not expect these models to encode any prosodic information given to their architecture, an expectation which was actually supported by a very low score on EmphAssess (18\%).

\subsection{Generalising the pipeline to S2S translation}

We now want to discuss how we can adapt our pipeline to S2ST capabilities. While most target languages can be evaluated directly using the existing pipeline, there are several considerations to remember. Firstly, it is essential to establish a validated topline. In other words, when introducing a new target language, we require validated translated utterances of the input English dataset in the desired language to have a topline in this target language. This process necessitates human validation, not only for the text translation, but also to either synthesise or record this translation with the correct emphasis, depending on the available resources. This new set of utterances can additionally serve as an input test set when we want to modify the source language to one other than English. 

Furthermore, we might want to modify or adapt some of the stages of the automatic evaluation pipeline in order to be better suited to the new language. For example, we have gathered evidence indicating that the emphasis classifier performs better when trained in the specific language it will be evaluated in. Thus, retraining it with emphasis data in the target language can prove advantageous, albeit demanding the corresponding larger dataset.

% \begin{figure*}[htpb]
%   \centering
%   \includegraphics[width=0.85\linewidth]{images/en-es_esclass_human.png}
%   \caption{Precision, recall and F1 scores on the English-to-Spanish models, using the Spanish Emphasis Classifier.}
%   \label{fig:en-es-results}
% \end{figure*}

We undertook a two-step process to modify our evaluation for English-to-Spanish translation. Firstly, external annotators translated the input sentences into Spanish, ensuring the inclusion of emphasis annotations. Subsequently, these translated sentences were synthesised into Spanish using our in-house TTS (Text-to-Speech) voices designed for Spanish, with a focus on retaining emphasis. Additionally, we adjusted the emphasis classifier to one specifically trained for Spanish as it yielded better results on Spanish data (see Appendix \ref{app:cross-lang-classifier}).

As depicted in the right panel of Figure \ref{fig:en-en-results}, the `topline,' which aligns the English input with the synthesised Spanish voices as the output, achieved a score of 58\%. While this result is reasonable, it notably lags behind the English topline. This decline may be attributed to various factors, including challenges in the synthesised voices, as we observed that our Spanish TTS voices do not emphasise as effectively as desired. Furthermore, issues in different stages of our automatic evaluation pipeline might contribute (for instance, the Spanish emphasis classifier's performance on spanish is not as optimal as its English counterpart on English data). Additionally, linguistic differences could play a role, with Spanish emphasis potentially being less prominent than in English or conveyed through alternative means, possibly paraphrastically in the text itself. Nonetheless, having this topline facilitates the comparison of other models and the assessment of their relative performance. Subsequently, we evaluated the Seamless M4T model \cite{barrault2023seamlessm4t} in its English-to-Spanish translation capability, which yielded an F1 score of 14\%. This result, akin to its English-to-English counterpart, suggests that the M4T model does not effectively capture emphasis.

\subsection{Human Evaluation}
To gauge human performance on the task, we conducted an evaluation with expert annotators. These annotators were presented with an utterance and its word-tokenised transcription, and were tasked with marking words they considered to be emphasised. Importantly, they were not obliged to mark any word as emphasised if they didn't perceive any. This evaluation was carried out on a subset of the data, incorporating both English and Spanish utterances, with native annotators for each language. 
% Precision, recall, and F1 scores are detailed in Table \ref{Tab:human-annotations}. 
Figure \ref{fig:en-en-results} shows precision, recall, and F1 scores for English-to-English and English-to-Spanish, respectively\footnote{For English-to-Spanish, the human topline is set using a subset of the Spanish utterances synthesized the Spanish topline}.
% For English-to-Spanish, the human topline is set using the same Spanish utterances from the Spanish topline.
These metrics were calculated by comparing the annotators' identification of emphasis against the `gold standard' annotation with which we synthesised the utterances.

Focusing first on the English dataset, the annotators achieved a commendable precision score of 86\%, although this was offset by a lower recall score (50\%). The lower recall could be attributed to annotators not perceiving emphasis in numerous sentences (Note: it is often harder to perceive emphasis in utterances taken out of their general, wider context); nonetheless, the high precision score is encouraging. Turning our attention to the Spanish dataset, both recall and precision scores were lower. This aligns with our hypothesis that the quality of voice synthesis in Spanish was not up to par - with the larger drop of recall compared to the topline could be explained by the Spanish emphasis classifier model picking up very subtle cues that are not obvious to the human ear. It may also suggest that the nuances of emphasis might be linguistically specific, thereby differing between English and Spanish.

% % Please add the following required packages to your document preamble:
% % \usepackage{booktabs}
% \begin{table}[]
% \begin{tabular}{@{}llll@{}}
% \toprule
% \textbf{} & Precision & Recall & F1 score \\ \midrule
% English   & 86\%      & 50\%   & 63\%     \\
% Spanish   & 58\%      & 41\%   & 48\%     \\ \bottomrule
% \end{tabular}
% \caption{Human annotated metrics on the English and Spanish synthesised utterances from the EmphAssess dataset. Each utterance was evaluated by 3 different annotators.}
% \label{Tab:human-annotations}
% \end{table}

\section{Conclusion}

We have introduced an evaluation framework for emphasis in speech-to-speech (S2S) models. This framework comprises an English dataset, an automated evaluation pipeline, and a results benchmark focusing on English-to-English and English-to-Spanish models. Crucially, our framework offers a generalisable approach applicable to other language pairs, the only major requirement being the acquisition of a relevant dataset to establish a reliable gold standard.

Additionally, we have open-sourced an emphasis-classification model that has been finetuned on English data. The model builds on a multilingual SSL architecture and has shown impressive accuracy in classifying emphasised speech in English on our dataset, along with reasonable performance in other languages (for further details, refer to the Appendix). The model's robustness in English makes it a plausible starting point for finetuning classifiers in other languages, potentially minimising the volume of data needed for training. Interestingly, the fact that the successful results were achieved without retraining the encoder, suggests that the inherent features in the original XLS-R model were adequate for emphasis classification.

There is an existing agenda for future research centring around the evaluation of prosody within SSL models. Firstly, on the subject of emphasis, we aim to scrutinise its functional role more closely—specifically, its ability to convey importance. We intend to investigate whether such a function is intrinsically represented within these models. Beyond emphasis, other aspects of prosody, such as turn-taking and speech grouping, merit attention. We are interested in determining whether these elements, too, are encoded within SSL models. Improved benchmarks and evaluations for these prosodic features could pave the way for the development of more expressive and nuanced models. 

To conclude, the EmphAssess benchmark sets a new standard for the evaluation of prosodic features in S2S models, offering both methodological contributions and actionable insights that could pave the way for more natural and effective machine-generated speech across various applications.

\section{Limitations}

While pioneering in its approach to evaluating emphasis in S2S models, our study encounters certain limitations. First, the emphasis classifier presented in this paper was made to be used with this exact dataset, and we recommend constraining its use to this particular use case (that is, with the presented benchmark and evaluation pipeline). Indeed, further testing is required to enhance its robustness and ensure its efficacy in detecting more nuanced forms of emphasis across other datasets.

Furthermore, the robustness of our evaluation process relies on the quality of multiple pipeline components, including Automatic Speech Recognition, forced alignment, and word-to-word alignment. Therefore, it is crucial to be mindful that errors could arise at various stages. Yet, the modular nature of the pipeline allows for continual improvements and assures that inter-model comparisons remain valid.

Another limitation of our work lies in the use of synthesised speech to create our dataset. While this approach provides a more controlled and consistent dataset—for instance, by enabling the synthesis of identical textual content with varying word emphases and voices—it may fail to capture the full range of characteristics found in natural speech. Consequently, this limitation could affect how well the benchmark results can be applied to practical use cases.

Lastly, our study is currently limited to binary categorisation of emphasis. Future endeavours could explore varying degrees of emphasis, although this would require more advanced models. For instance, capturing subtle differences in emphasis between the input and output of an S2S system could be a valuable addition to this line of research.

% Requires by the new ARR template. I moved the limitations from above here

\ifanonymous
    % Nothing
\else
    \section*{Acknowledgements}
    ED in his EHESS capacity has been funded by the Agence Nationale pour la Recherche (ANR-17-EURE-0017 Frontcog, ANR-10-IDEX-0001-02 PSL*, ANR-19-P3IA-0001 PRAIRIE 3IA Institute) and a grant from CIFAR (Learning in Machines and Brains).
\fi

% \bibliography{anthology,custom}
\bibliography{custom}

\newpage
% \onecolumn
\appendix

\section{Cross-language generalisation in the classifier }\label{app:cross-lang-classifier}

Using a Spanish company-internal variant of the Expresso dataset, we trained and tested the classifier on Spanish data in an identical manner to our approach with English.  We should however note that the version of the data we had was of lesser recording quality than the English one. 

The classifier's outcomes when evaluated on both the English and Spanish train sets are presented in Table \ref{Tab:EmphClassCrossLang}. The most important observation from the results is the classifier's superior performance when trained and tested on the same language. Cross-language assessments, especially from English-trained models tested on Spanish data, manifested a decline in performance.  Nevertheless, despite the noted challenges, the results demonstrate that the classifier is able to detect emphasis, even across languages. It is also worth that the Spanish dataset was of considerably lower quality than the English one and is just used here for demonstration purposes. It is plausible that this quality might have affected the model's performance. Therefore, a more definitive assessment of its cross-language generalisation potential would necessitate testing on datasets of other languages, ideally of comparable quality to the English version.

We also extended the evaluation of the English and Spanish emphasis classifiers to additional languages, using internal datasets to compile test sets mirroring the structure of the English ones, each featuring 2 to 3 speakers. These are summarised in Table \ref{Tab:EmphClassCrossLang}. Intriguingly, the Spanish classifier outperformed across all tested languages, a finding readily attributable to linguistic similarities in the case of Italian, French, and Portuguese, but less so for Vietnamese.  Furthermore, in some instances, performance on non-native test sets was on par with, or even surpassed, native datasets; for example, a word-level F1 score of 84.4\% was achieved on the Portuguese test set. These observations imply the feasibility of applying classifiers to languages they were not specifically trained on, particularly when sufficient training data is lacking, and suggest the merit in experimenting with classifiers based on different languages. Additional results could potentially advocate for the benefits of multi-language training approaches. An additional point of interest arises from the performance of the Vietnamese test sets. Vietnam's tonal nature, which distinctly shapes its emphasis patterns, ostensibly diverges from the prosodic systems used in Romance and Germanic languages. Despite these fundamental differences, the fact that the Spanish-trained classifier achieved commendable results with Vietnamese indicates that it may be recognising universal features of emphasis that transcend language-specific prosodic systems. 

% \textcolor{red}{To-do : evaluate other languages from mExpresso and add scores. Discuss tonal Vietnamese. Discuss also the fact that generally better with Portuguese data (that we can't share). Overall better on Spanish data, potentially because of distance (except for Prtuguese. Discuss if more at recall or prec level}

\begin{table*}[b]
    \centering
    \begin{tabular}{llllllll} 
    \toprule
         &  &  \multicolumn{3}{l}{\textbf{Frame-level metrics (\%)}}&  \multicolumn{3}{c}{\textbf{Word-level metrics (\%)}}\\ 
         Test data&  Train data&  \textbf{F1 score}& \textbf{ Precision}& \textbf{ Recall}&  \textbf{F1 score}& \textbf{ Precision}& \textbf{Recall}\\ 
         \midrule
         \rowcolor{gray!20}
         English&  English&   \textbf{75.52}&   77.48&   76.9&   \textbf{78.4}&   78.96&  79.46\\ 
        English&  Spanish&  67.36&  68.74&  71.95&  68.66&  66.73& 75.21\\ 
         \midrule
         Spanish&  English&  55.75&  60.82&  55.16&  56.14&  56.93& 57.92\\ 
         \rowcolor{gray!20}
         Spanish&  Spanish&  \textbf{72.52}&  73.26&  75.12&  \textbf{73.92}&  74.21& 76.32\\ 
         \midrule
         Vietnamese&  English&  61.65&  68.98&  61.51&  64.59&  70.63& 63.7\\ 
         Vietnamese&  Spanish&  \textbf{71.21}&  71.82&  76.32&  \textbf{75.48}&  77.69& 78.2\\ 
         \midrule
         Italian&  English&   56.79&   70.61&   52.86&   56.12&   57.18&  57.61\\ 
         Italian&  Spanish&  \textbf{64.72}&  72.64&  63.46&  \textbf{67.81}&  68.42& 70.41\\ 
         \midrule
 French& English& 60.18& 62.81& 63.31& 65.08& 65.85&67.07\\ 
 French& Spanish& \textbf{62.50}& 63.09& 68.05& \textbf{68.17}& 67.64&72.41\\ 
 \midrule
 Portuguese& English&  71.84&  83.56&  68.41&  72.86&  73.17& 74.69\\ 
 Portuguese& Spanish& \textbf{79.84}& 82.93& 80.08& \textbf{84.4}& 84.15&87.1\\ 
 \bottomrule
    \end{tabular}
    \caption{Performance metrics of the emphasis classifier across multiple languages, benchmarked using F1 score, precision, and recall. The classifier is trained either on English or Spanish data sets. Rows highlighted in grey represent instances where the training and test data languages are identical.}
    \label{Tab:EmphClassCrossLang}

\end{table*}

\end{document}